\documentclass[a4paper, 11pt]{article}
\usepackage{geometry}
\usepackage{setspace}
\usepackage{multirow}
\usepackage{graphics}
\usepackage{graphicx}
\usepackage{amsfonts}
\usepackage{amsmath}
\usepackage{amssymb}
\usepackage{authblk}
\usepackage{times}
\usepackage{url}
\usepackage[utf8]{inputenc}
\usepackage[T1]{fontenc}
\usepackage[version=4]{mhchem}
\usepackage[numbers, sort&compress, square]{natbib}

\title{MLSolv-A: A Novel Machine Learning-Based Prediction of Solvation Free Energies from Pairwise Atomistic Interactions}
\author{Hyuntae Lim\thanks{ht0620@snu.ac.kr}}
\author{YounJoon Jung\thanks{yjjung@snu.ac.kr}}
\affil{Department of Chemistry, Seoul National University, Seoul 08826, Korea}
\date{\today}

\begin{document}

\maketitle

\begin{abstract}
    Recent advances in machine learning and their applications have lead to the development of diverse structure-property relationship models for crucial chemical properties, and the solvation free energy is one of them.
    Here, we introduce a novel ML-based solvation model, which calculates the solvation energy from pairwise atomistic interactions.
    The novelty of the proposed model consists of a simple architecture: two encoding functions extract atomic feature vectors from the given chemical structure, while the inner product between two atomistic features calculates their interactions.
    The results on 6,493 experimental measurements achieve outstanding performance and transferability for enlarging training data due to its solvent-non-specific nature.
    Analysis of the interaction map shows there is a great potential that our model reproduces group contributions on the solvation energy, which makes us believe that the model not only provides the predicted target property but also gives us more detailed physicochemical insights.
\end{abstract}


\section{Introduction}\label{Introduction}

The importance of solvation or hydration mechanism and accompanying free energy change has made various \emph{in silico} calculation methods for the solvation energy one of the most important application in computational chemistry\cite{klamt_cosmo:_1993,delaney_esol:_2004,tomasi_quantum_2005,cramer_universal_2008,marenich_universal_2009,klamt_cosmo-rs:_2010,shivakumar_prediction_2010,chong_atomic_2011,mennucci_polarizable_2012,lim_delfos:_2019,marenich_generalized_2013,sato_modern_2013,konig_predicting_2014,mobley_freesolv:_2014,klamt_calculation_2015,skyner_review_2015,zhang_force_2015,harder_opls3:_2016,coley_convolutional_2017,duarte_ramos_matos_approaches_2017,wu_moleculenet:_2018,borhani_hybrid_2019,ringe_function-space-based_2016,hille_generalized_2019,chong_interaction_2014}.
The solvation free energy directly influences many chemical properties in solvated phases and plays a dominant role in various chemical reactions: drug delivery\cite{delaney_esol:_2004,skyner_review_2015,harder_opls3:_2016,popova_deep_2018}, organic synthesis\cite{reichardt_solvents_2010}, electrochemical redox reactions\cite{takeda_electron-deficient_2016,park_high-efficiency_2017,allam_application_2018,kim_biological_2019}, etc.

The atomistic computer simulation approaches for the solvent and the solute molecules directly offer the microscopic structure of the solvation shell, which surrounds the solutes molecule\cite{shivakumar_prediction_2010,chong_atomic_2011,konig_predicting_2014,zhang_force_2015,harder_opls3:_2016,jia_calculations_2016}.
The solvation shell structure could provide us detailed physicochemical information like microscopic mechanisms on solvation or the interplay between the solvent and the solute molecules when we use an appropriate force field and molecular dynamics parameters.
However, those \emph{explicit solvation} methods we stated above need an extensive amount of numerical calculations since we have to simulate each individual molecule in the solvated system.
The practical problems on the explicit solvation model restrict its applications to classical molecular mechanics simulations\cite{shivakumar_prediction_2010,chong_atomic_2011,zhang_force_2015} or a limited number of QM/MM approaches\cite{konig_predicting_2014,jia_calculations_2016}.

For classical mechanics approaches for macromolecules or calculations for small compounds at quantum-mechanical level,
the idea of \emph{implicit solvation} enables us to calculate solvation energy with feasible time and computational costs when one considers a given solvent as a continuous and isotropic medium in the Poisson-Boltzmann equation\cite{klamt_cosmo:_1993,tomasi_quantum_2005,cramer_universal_2008,marenich_universal_2009,klamt_cosmo-rs:_2010,mennucci_polarizable_2012,marenich_generalized_2013,klamt_calculation_2015,ringe_function-space-based_2016,hille_generalized_2019}.
Many theoretical advances have been introduced to construct the continuum solvation model, which involves parameterized solvent properties: the polarizable continuum model (PCM)\cite{mennucci_polarizable_2012}, the conductor-like screening model (COSMO)\cite{klamt_cosmo:_1993} and its variations\cite{klamt_cosmo-rs:_2010,klamt_cosmo_2018}, generalized Born approximations like solvation model based on density (SMD)\cite{marenich_universal_2009} or solvation model 6, 8, 12, etc. (SMx)\cite{cramer_universal_2008,marenich_generalized_2013}.

The structure-property relationship (SPR) is rather a new approach,
which predicts the solvation free energy with a completely different point of view when compared to computer simulation approaches with precisely defined theoretical backgrounds\cite{tropsha_best_2010,cherkasov_qsar_2014}.
Although we may not fully expect to obtain detailed chemical or physical insights other than the target property,
since the SPR is a regression analysis in its intrinsic nature, it has demonstrated great potentials in terms of transferability and outstanding computational efficiency\cite{tropsha_best_2010,cherkasov_qsar_2014,wu_moleculenet:_2018}.
Recent successes in the machine learning (ML) technique\cite{schmidhuber_deep_2015} and their implementations in computational chemistry\cite{wu_moleculenet:_2018,butler_machine_2018} are currently promoting broad applications of SPR in numerous chemical studies\cite{delaney_esol:_2004,kearnes_molecular_2016,coley_convolutional_2017,gilmer_neural_2017,schutt_quantum-chemical_2017,smith_ani-1:_2017,behler_first_2017,allam_application_2018,popova_deep_2018,ryu_deeply_2018,schutt_schnet_2018,sifain_discovering_2018,borhani_hybrid_2019,ryu_bayesian_2019,schutt_schnetpack:_2019,winter_learning_2019,zhang_prediction_2019}.
Those studies proved that ML guarantees faster calculations than computer simulations and more precise estimations than traditional SPR estimations;
a decent number of models showed accuracies comparable to \emph{ab initio} solvation models in the aqueous system\cite{wu_moleculenet:_2018}.

We introduced a novel artificial neural-network-based ML solvation model called \emph{Delfos} that predicts free energies of solvation for generic organic solvents in the previous work\cite{lim_delfos:_2019}.
The model not only has a great potential of showing an accuracy comparable to the state-of-the-art computational chemistry methods\cite{marenich_generalized_2013,klamt_calculation_2015} but offers information of which substructures play a dominant role in the solvation process.
In the present work, we propose a novel approach to ML model for the solvation energy estimation called \emph{MLSolv-A}, which is based on the group-contribution method.
The key idea of the proposed model is the calculation of pairwise atomic interactions by mapping them into inner products of atomic feature vectors, while each encoder network for the solvent and the solute extracts such atomic features.
We believe the proposed approach can be a powerful tool for understanding the solvation process, not only a separate or alternative prediction method, but also can strengthen various solvation models via computer simulations.

The outline of the rest of the present paper is as follows:
in section \ref{Methods}, we introduce the theoretical backgrounds  of ML techniques we use and the overall architecture of our proposed model.
Section \ref{Results} quantifies the model's prediction performance with 6,594 data points, mainly focused on pairwise atomic interactions and corresponding group contributions on the solvation free energy.
In the last section of the present paper, we summarize and conclude our work.


\section{Methods}\label{Methods}

In the proposed model, the linear regression task between the given chemical structures of the solvent and solute molecules and their free energy of solvation starts with the atomistic vector representations\cite{jaeger_mol2vec:_2018,lim_delfos:_2019} of the given solvent molecule, consisting of $\mathbf{x}_{\alpha}$'s and solute molecule consisting of $\mathbf{y}_{\gamma}$'s, where $\alpha$ and $\gamma$ are the atom indices. The entire molecular structure is now can be expressed as a sequence of vectors or a matrix:
\begin{subequations}
    \begin{align}
        \mathbf{X} = \left\{ \mathbf{x}_{\alpha} \right\},
        \\
        \mathbf{Y} = \left\{ \mathbf{y}_{\gamma} \right\},
    \end{align}
    \label{eqn:embedding}
\end{subequations}
where $\mathbf{x}_{\alpha}$ and $\mathbf{y}_{\gamma}$ are the $\alpha$-th row of $\mathbf{X}$ and the $\gamma$-th row of $\mathbf{Y}$, respectively.
Then the encoder function learns of their chemical structures and extracts feature matrices for the solvent $\mathbf{P}$ and the solute $\mathbf{Q}$,
\begin{subequations}
    \begin{align}
        \mathbf{P} = \left\{ \mathbf{p}_{\alpha} \right\} = \mathrm{Encoder} (\mathbf{X}),
        \\
        \mathbf{Q} = \left\{ \mathbf{q}_{\gamma} \right\} = \mathrm{Encoder} (\mathbf{Y}).
    \end{align}
\end{subequations}
Rows of $\mathbf{P}$ and $\mathbf{Q}$, $\mathbf{p}_{\alpha}$ and $\mathbf{q}_{\gamma}$ involve atomistic chemical features of atoms $\alpha$ and $\gamma$, which are directly related to the target property, which is the solvation free energy in our model.
We now calculate the un-normalized attention score (or \emph{chemical similarity}) between the atoms $\alpha$ and $\gamma$ with the Luong's dot-product attention\cite{luong_effective_2015}:
\begin{equation}
    I_{\alpha \gamma} = - \mathbf{p}_{\alpha} \cdot \mathbf{q}_{\gamma}.
    \label{eqn:atomic_interaction}
\end{equation}
Since our target quantity is the free energies of solvation, we expect such chemical similarity $I_{\alpha \gamma}$ to correspond well to atomistic interactions between $\alpha$ and $\gamma$, which involves both the energetic and the entropic contributions.
Eventually, the free energy of solvation of the given solvent-solute pair, which is the final regression target, is expressed as a simple summation of atomistic interactions:
\begin{equation}
    \Delta G_{sol}^{\circ} = \sum_{\alpha\gamma} I_{\alpha\gamma}.
    \label{eqn:summation_interaction}
\end{equation}
Certainly, one can also calculate the free energies of solvation from two molecular feature vectors, which represent the solvent properties $\mathbf{u}$ and the solute properties $\mathbf{v}$, respectively:
\begin{equation}
    \Delta G_{sol}^{\circ} = \mathbf{u} \cdot \mathbf{v}
    = \left( \sum_{\alpha} \mathbf{p}_{\alpha} \right) \cdot \left( \sum_{\alpha} \mathbf{q}_{\alpha} \right).
\end{equation}
The inner-product relation between molecular feature vectors $\mathbf{u}$ and $\mathbf{v}$ has a formal analogy with the solvent-gas partition coefficient calculation method via the solvation descriptor approach\cite{stolov_enthalpies_2017,sedov_abraham_2018}.

\begin{figure}
    \centering
    \includegraphics[width = 10cm]{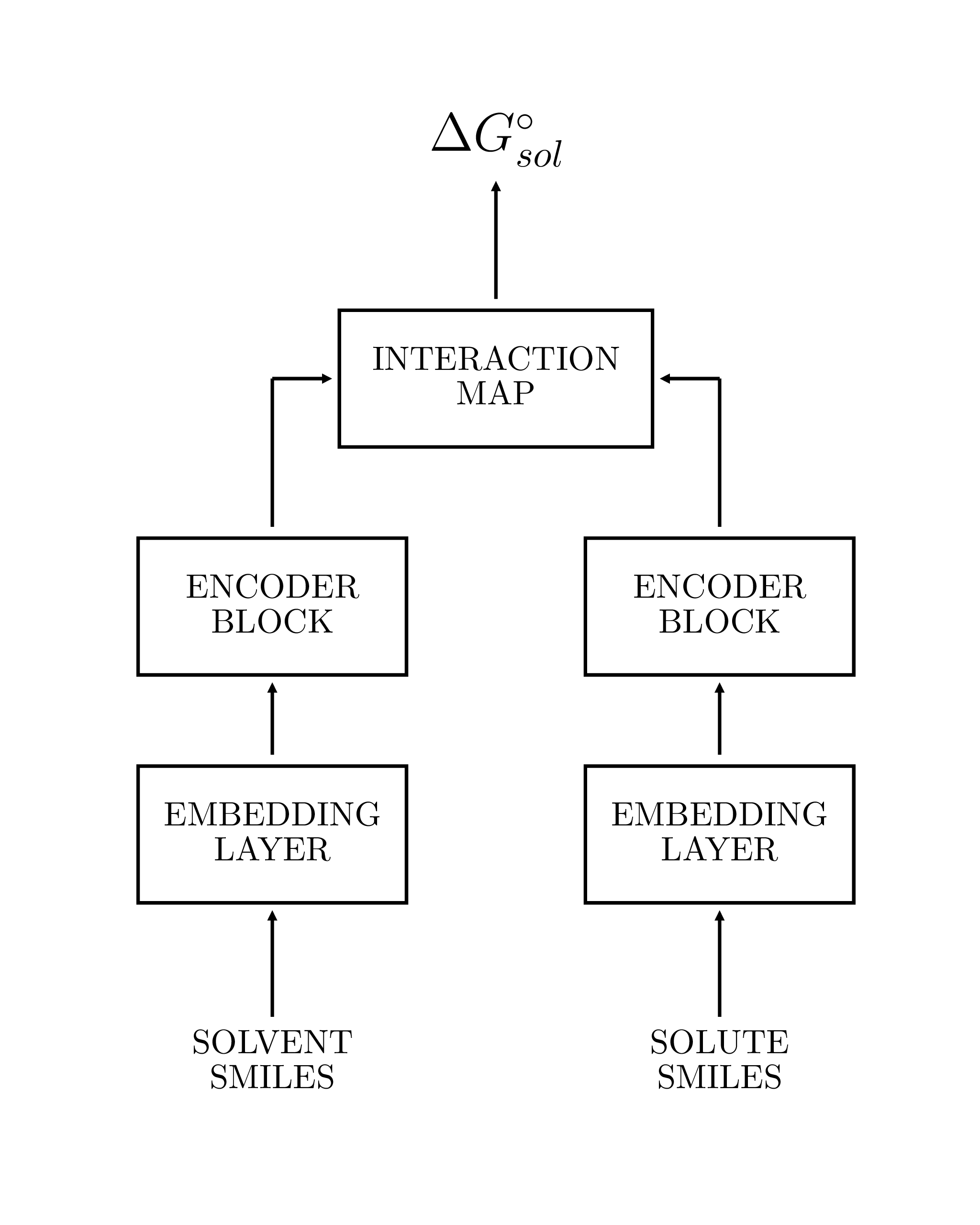}
    \caption{Schematic illustration of the architecture of MLSolv-A. Each encoder network extracts atomistic feature vectors given pre-trained vector representations, and the interaction map calculates pairwise atomistic interactions from Luong's dot-product attention\cite{luong_effective_2015}.}
    \label{fig:architecture}
\end{figure}

We choose and compare two different neural network models in order to encode the input molecular structure and extract important structural or chemical features which are strongly related to solvation behavior: one is the bidirectional language model (BiLM)\cite{peters_deep_2018}, based on the recurrent neural network (RNN), the other is the graph convolutional neural network (GCN)\cite{kipf_semi-supervised_2017} which explicitly handles the connectivity (bonding) between atoms with the adjacency matrix (more detailed description available in the \textbf{supporting information}).
Figure \ref{fig:architecture} illustrates an overview on the architecture of the proposed ML solvation model.

\section{Results and Discussions}\label{Results}

For the training and test tasks of the proposed neural network, we prepare 6,594 experimental measures of free energies of solvation for 952 organic solvents and 147 organic solutes, including some inert gases.
642 experimentally measured values of the free energy of hydration are taken from the FreeSolv database\cite{mobley_mobleylabfreesolv_2018,mobley_freesolv:_2014},
and 5,952 data points for non-aqueous solvents are collected with the Solv@TUM database version 1.0\cite{christoph_1_solvtum_2018,hille_generalized_2019,stolov_enthalpies_2017,sedov_abraham_2018}, which is available at https://mediatum.ub.tum.de/1452571.
Compounds in the dataset involves 10 kinds of atoms, which are commonly used in organic chemistry: hydrogen (H), carbon (C), oxygen (O), sulfur (S), nitrogen (N), phosphorus (P), fluorine (F), chlorine (Cl), bromine (Br), and iodine (I). The maximum heavy-atom count is 28 for the solute molecules and 18 for the solvent molecules.

At the very first stage, we perform the skip-gram pre-training process for 10,229,472 organic compounds, which are collected from the ZINC15 database\cite{sterling_zinc_2015}, with Gensim 3.8.1 and Mol2Vec skip-gram model to construct the 128-dimensional embedding lookup table\cite{jaeger_mol2vec:_2018}.
For the implementation of the neural network model, we mainly use the Tensorflow 2.0 and Keras 2.3.1 frameworks\cite{martin_abadi_tensorflow:_2015}.
Each model has L2 regularization to prevent excessive changes on weights and to minimize the variance, and uses the RMSprop algorithm with $10^{-3}$ of learning rate and $\rho = 0.9$ for optimizing its loss function, the mean squared error (MSE).
The selection of the optimized model for the target property is realized by an extensive grid-search task for tuning model hyperparameters.

\begin{figure}
    \centering
    \includegraphics[width = 15cm]{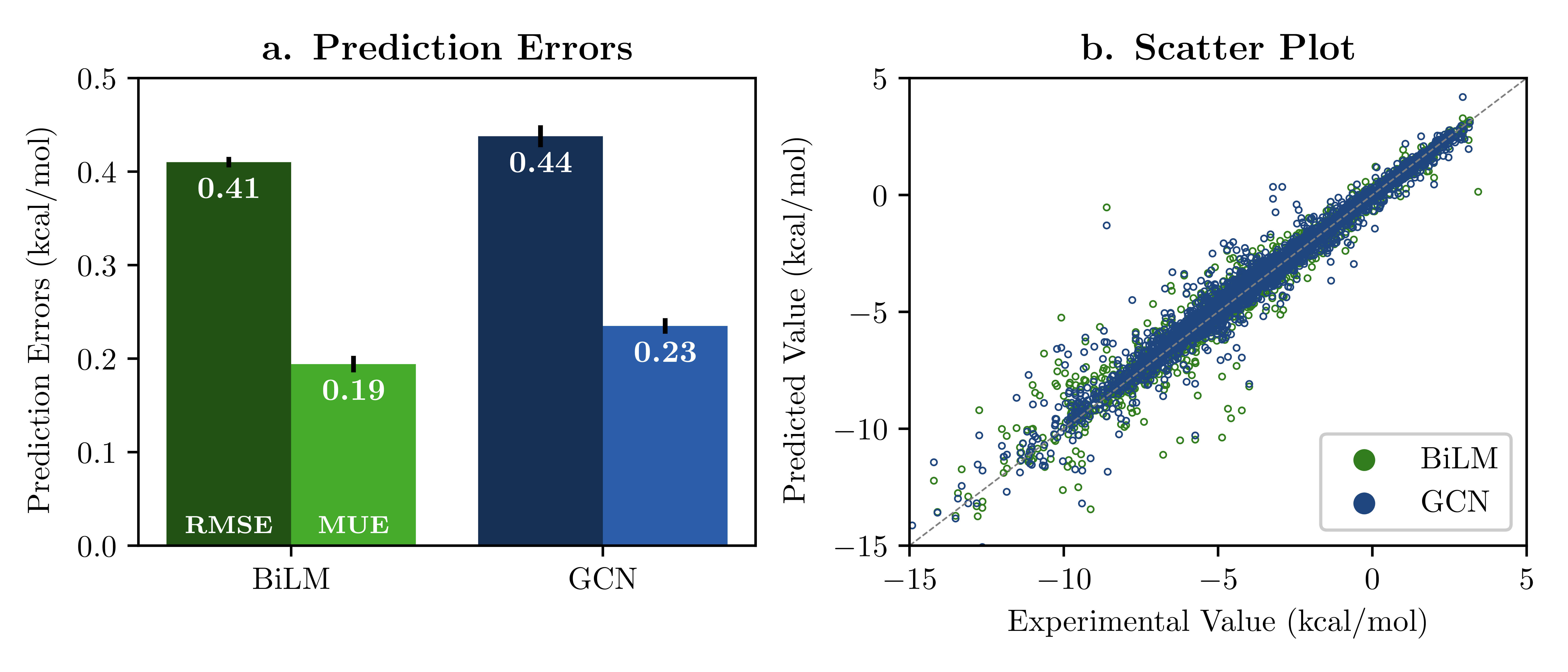}
    \caption{(a) Prediction erros for two models in $\mathrm{kcal/mol}$, taken from 5-fold cross validation results. (b) Scatter plot between the experimental value and ML the ML predicted value. Black circles denote the BiLM model while the GCN results are shown in gray diamonds.}
    \label{fig:cv}
\end{figure}

We employ 5-fold cross-validation to evaluate the prediction accuracy of the chosen model\cite{pedregosa_scikit-learn:_2011};
the entire dataset is randomly split into five uniform-sized subsets, and we iteratively choose one of the subsets as a test set, and the training run uses the remaining 4 subsets.
Consequentially, a 5-fold CV task performs 5 independent training and test runs, and relative sizes of the training and test sets are 8 to 2.
To minimize the variation of results from CV tasks, we take averages for all results over 9 independent random CVs, split from different random states.
The results for test run using 5-fold CV tasks for the optimized models are shown in Fig. \ref{fig:cv}.
We found that the BiLM encoder with the LSTM layer performs slightly better than the GCN encoder, although their differences are not pronounced:
the mean unsigned prediction error (MUE) for the BiLM/LSTM encoder model is $0.19 ~\mathrm{kcal/mol}$, while the GCN model results in $0.23 ~\mathrm{kcal/mol}$.
Both MUE values show that the our proposed mechanism works well and guarantees excellent prediction accuracies for well-trained chemical structures.
Moreover, projecting from our results based on a simple version of the graph-based neural network as the encoder, we expect the GCN-based model to perform better than a simple graph-based embedding model, or more progressed version of graph neural networks to perform even better for chemical structures: such as the messege-passing neural network (MPNN)\cite{gilmer_neural_2017}, the deep tensor neural network (DTNN)\cite{schutt_quantum-chemical_2017}, and so on.

\begin{figure}
    \centering
    \includegraphics[width = 15cm]{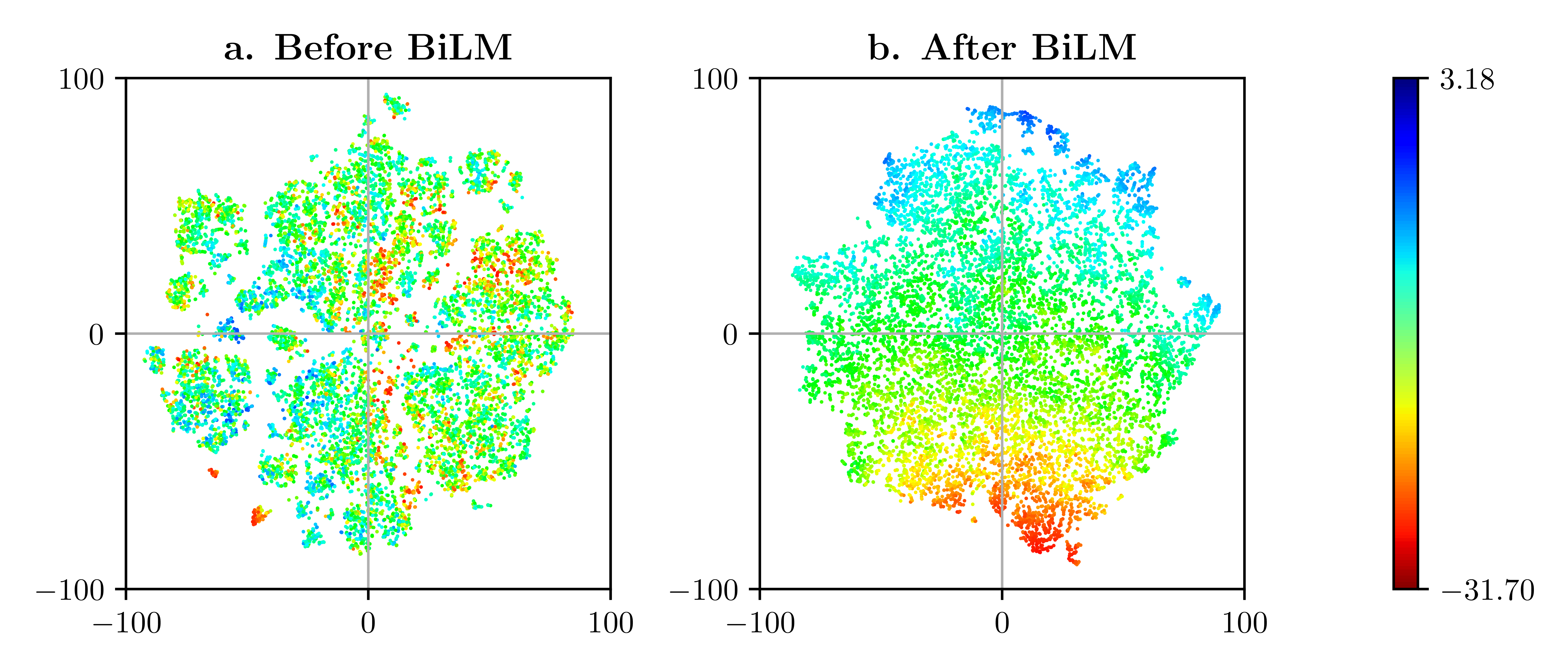}
    \caption{2-dimensional visualizations on (a) the pre-trained vector from the skip-gram model $\sum_{\gamma} \mathbf{y}_{\gamma}$ and (b) the extracted molecular feature vector $\mathbf{v}$ for 15,432 solutes. We reduce the dimension of each vector with the t-SNE algorithm. The color representation denotes the hydration energy of each point.}
    \label{fig:tsne}
\end{figure}

Figure \ref{fig:tsne} presents t-SNE visualizations for pre-trained solute vectors $\mathbf{y}$ and encoded molecular feature $\mathbf{v}$\cite{ryu_deeply_2018}, which confirms whether or not the proposed neural network architecture works as we designed.
Color codes denote predicted hydration free energies for 15,432 points, whose structures are randomly taken from the ZINC15\cite{sterling_zinc_2015}; red dots correpond to low hydration free energy case, while the blue to high hydration free energy cases.
The correlation between molecular features and predicted free energies provides a clear evidence that the model architecture indeed extracts geometrical correlations and calculates the free energy accurately.
Meanwhile, the pre-trained solute vectors from the skip-gram embedding model exhibit only weak correlations.

\begin{figure}
    \centering
    \includegraphics[width = 10cm]{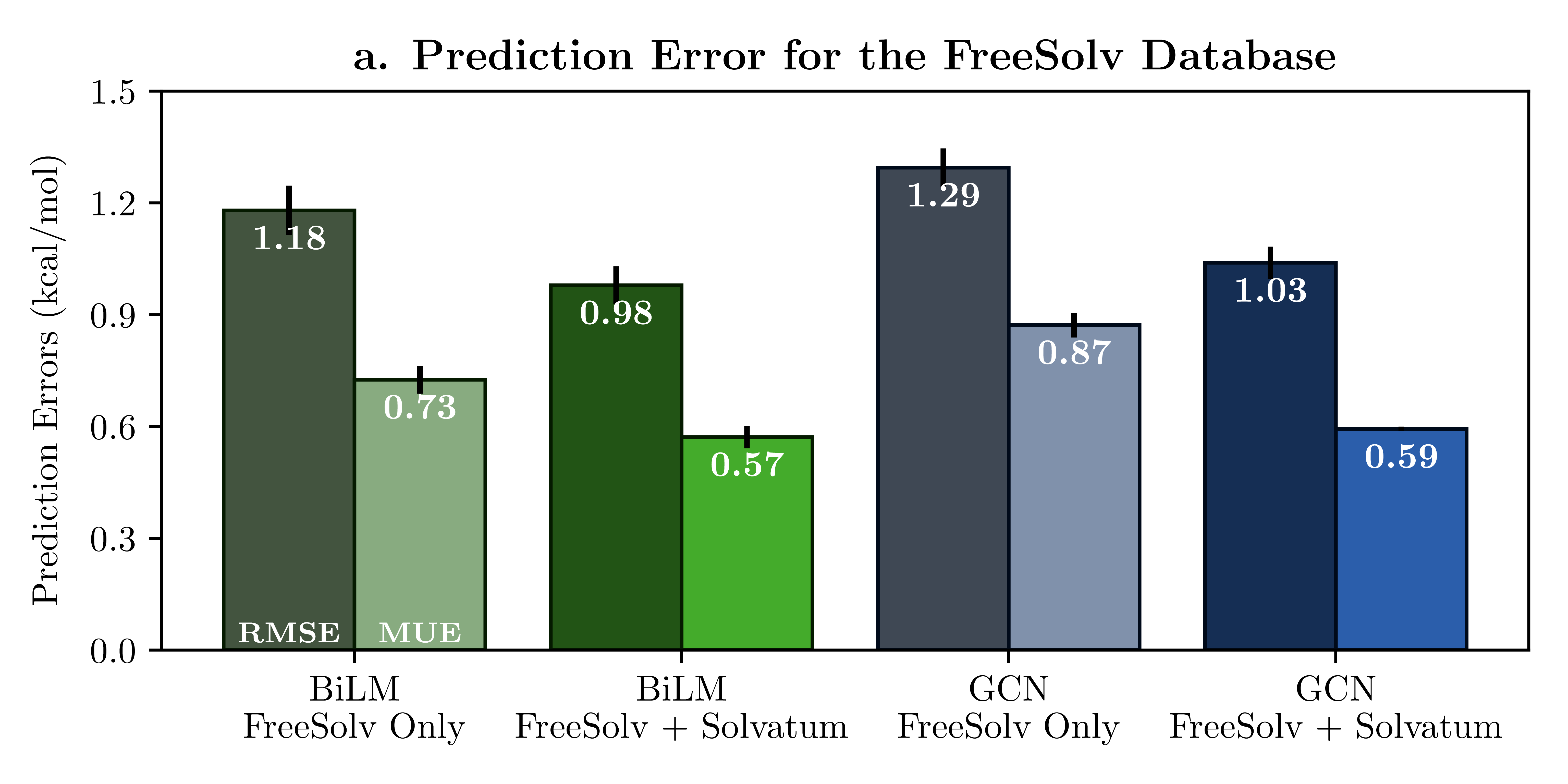}
    \includegraphics[width = 10cm]{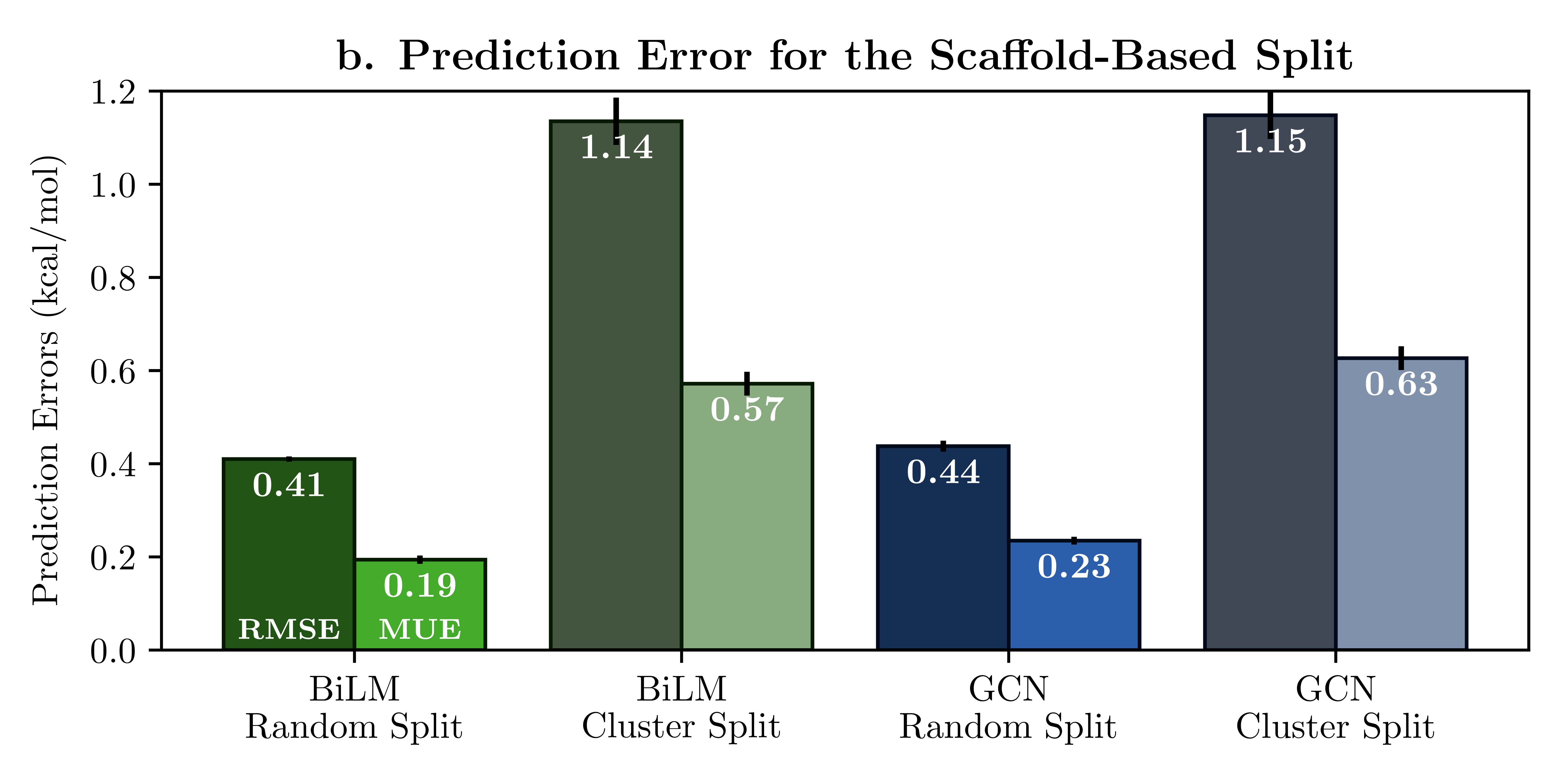}
    \caption{(a) CV-results for FreeSolv hydration energies with two different training dataset selection. Deep-colored boxes denote CV results with the augmented dataset with the Solv@TUM database.
    (b)Comparison between CV results with the random-split and the scaffold-based split (or cluster split).}
    \label{fig:hydration}
\end{figure}

Since our proposed neural network model is a solvent-non-specific one that considers both the solvent and the solute structures as separated inputs, it has a distinct character when compared to other solvent-specific ML solvation models.
The model can train with the molecular structure of a single solute repeatedly when the solute has multiple solvation energy data points for different kinds of solvents\cite{lim_delfos:_2019};
this logic is also valid for a single solvent.
Therefore, one of the most useful advantages of our model is that we can easily enlarge the dataset for training, even in the scenario that we want to predict solvation free energies for a specific solvent.
Figure \ref{fig:hydration} shows 5-fold CV results for 642 hydration free energies (FreeSolv) from both the BiLM and the GCN models, in two different situations.
One uses only the FreeSolv\cite{mobley_freesolv:_2014,mobley_mobleylabfreesolv_2018} database for train and test tasks, and the other additionally uses the Solv@TUM\cite{christoph_1_solvtum_2018,hille_generalized_2019}.
Although the Solv@TUM database only involves non-aqueous data points, it enhances each model's accuracy by about 20\% (BiLM) to 30\% (GCN) in terms of MUE.
Those results imply that there are possible applications of the transfer learning to other solvation-related properties, like aqueous solubilities\cite{delaney_esol:_2004} or octanol-water partition coefficients.

However, in some other situations, one may be concerned that the repetitive training for a single compound may make the model tends to overfit, and they could weaken predictivity for the structurally new compound, which is considered as an extrapolation.
We investigate the model's predictivity for extrapolation situations with the \emph{scaffold-based} split\cite{winter_learning_2019,gilmer_neural_2017,lim_delfos:_2019}.
Instead of the ordinary K-fold CV task with the random and uniform split method, the K-means clustering algorithm builds each fold with the MACCS substructural fingerprint\cite{winter_learning_2019}.
One can simulate an extreme extrapolation situation through CV tasks over the clustered folds.
As shown in Fig. \ref{fig:hydration}, albeit the scaffold-based split degrades MUEs by a factor of three, they are still within an acceptable error range $\sim 0.6 ~\mathrm{kcal/mol}$, given chemical accuracy $1.0 ~\mathrm{kcal/mol}$.
Furthermore, we do not see any evidence that our model tends to overfit more than other solvent-specific models\cite{gilmer_neural_2017,winter_learning_2019}.

\begin{figure}
    \centering
    \includegraphics[width = 9cm]{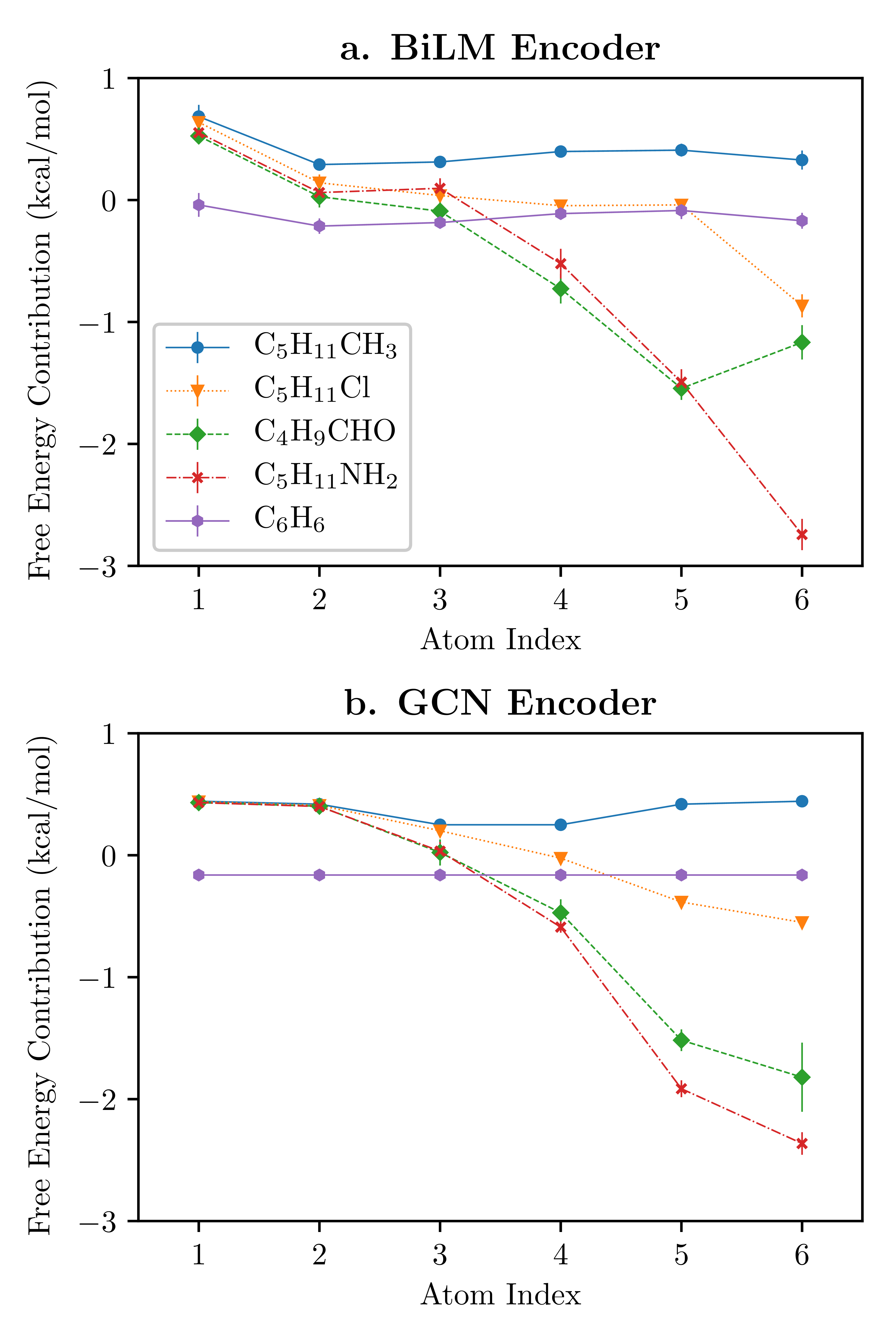}
    \caption{ML-calculated atomistic group contributions for five small organic compounds which have six heavy atoms (excluding hydrogens). The atom index starts from the leftmost of the given molecule and only counts heavy atoms.}
    \label{fig:group}
\end{figure}

Although we showed that the proposed NN model guarantees an excellent predictivity for solvation energies of various solute and solvent pairs,
the main objective of the present study is to obtain the solvation free energy as the sum of decomposed inter-atomic interactions, as we described in Eqs. \ref{eqn:atomic_interaction} and \ref{eqn:summation_interaction}.
In order to verify the feasibility of the the model's solvation energy estimation to decompose into group contributions,
we define the sum of atomic interactions $\mathbf{I}_{\alpha \gamma}$ over the solvent indices $\gamma$ as the group contributions of the $\alpha$-th solute atom:
\begin{equation}
    \mathbf{I}_{\alpha} = \sum_{\gamma} \mathbf{I}_{\alpha \gamma}
    \label{eqn:group_contribution}.
\end{equation}
Figure \ref{fig:group} shows hydration free energy contributions for five small organic solutes which have six heavy atoms: n-hexane, 1-chloropentane, pentaldehyde, 1-aminopentane, and benzene.
As shown in Fig. \ref{fig:group}, both the BiLM and the GCN model exhibit a similar tendency in group contributions;
the model estimates that atomic interactions between the solute atoms and water increases near the hydrophilic groups.
It is obvious that each atom in benzene must have identical contributions to the free energy,
but the results in Fig. \ref{fig:group} clearly shows that the BiLM model makes faulty predictions while the GCN model works well as expected.
We believe that this malfunctioning of the BiLM model originates from the sequential nature of the recurrent neural network.
Since the RNN considers the input molecule is just a simple sequence of atomic vectors and there are no explicit statements that involve bonding information, the model could not be aware of the cyclic shape of the input compound\cite{kearnes_molecular_2016,popova_deep_2018}.
We conclude that it is inevitable to use explicitly bond (or connectivity) information when one constructs a group-contribution based ML model, albeit the RNN-based model predicts well in terms of their sum.

\section{Conclusions}\label{Conclusions}

In summary, we introduced a novel approach for the ML-based solvation energy prediction, which has a great potential to provide physicochemical insights on the solvation process.
The novelty of our neural network model is that the model does not involve the perceptron networks for readout of encoded features and estimation of the target property.
Alternatively, we designed the model such that it is possible to calculate pairwise atomic interactions from inner products of atomistic feature vectors\cite{luong_effective_2015}.
As a result, the model produces the solvation free energy from the group-contribution based prediction.

We quantified the proposed model's prediction performances for 6,493 experimental data points of solvation energies, which were taken from the FreeSolv\cite{mobley_freesolv:_2014,mobley_mobleylabfreesolv_2018} and Solv@TUM database\cite{christoph_1_solvtum_2018,hille_generalized_2019}.
We found a significant geometrical correlation between molecular feature vectors and predicted properties, which confirms that the proposed model successfully extracts chemical properties and maps into the vector representations.
 is actually working as we designed.
The estimated prediction MUEs from K-fold CV are $0.19 ~\mathrm{kcal/mol}$ for the BiLM encoder and $0.23 ~\mathrm{kcal/mol}$ for the GCN model, respectively.

The K-fold CV results from the scaffold-based split\cite{winter_learning_2019} showed the prediction accuracy decreases by a factor of three in extreme extrapolation situations, but they still exhibit moderate performances, which were $0.60 ~\mathrm{kcal/mol}$.
Moreover, we found that the solvent-non-specific structure of the proposed model is appropriate for enlarging the dataset size,
i.e., experimental data points for a particular solvent are transferable to other solvents;
we conclude that this transferability is the reason for our model's outstanding predictivity\cite{lim_delfos:_2019}.

Finally, we examined pairwise atomic interactions that are obtained from the interaction map and found a clear tendency between hydrophilic groups and their contributions to the hydration free energy.
We believe that our model can provide detailed information on the solvation mechanism, not only the predicted value of the target property.

\section{Conflicts of interest}
There are no conflicts to declare.

\bibliographystyle{achemso}
\bibliography{references}

\end{document}


\maketitle

\setstretch{1.5}

\section*{Detailed Model Description}

The detailed mathematical expressions of the bidirectional language model (BiLM) are as follows\cite{peters_deep_2018}:
\begin{subequations}
    \begin{align}
        \overrightarrow{\mathbf{H}}^{(i + 1)} =
        \overrightarrow{\mathrm{RNN}} (\overrightarrow{\mathbf{H}}^{(i)}),
        \\
        \overleftarrow{\mathbf{H}}^{(i + 1)} =
        \overleftarrow{\mathrm{RNN}} (\overleftarrow{\mathbf{H}}^{(i)}).
    \end{align}
    \label{eqn:group:bilm}
\end{subequations}
In Eqn. \ref{eqn:group:bilm}, the right-headed arrow in $\overrightarrow{\mathrm{RNN}}$ denotes a forward-directed recurrent unit which propagates from the leftmost of the sequence to the rightmost one.
The BiLM also involves the backward-directed recurrent neural network ($\overleftarrow{\mathrm{RNN}}$) and it propagates from the rightmost to the leftmost.
The superscript $(i)$ in hidden layers $\mathbf{H}^{(i)}$ denotes the position at the stacked configuration: at the first layer, both forward-directed and backward-directed RNN share the pre-trained sequence $\mathbf{X}$ as an input, $\overrightarrow{\mathbf{H}}^{(0)} = \overleftarrow{\mathbf{H}}^{(0)} = \mathbf{X}$.
In addition, use of more improved versions of RNNs, e.g. the gated recurrent unit (GRU)\cite{chung_empirical_2014} or the long-short term memory (LSTM)\cite{hochreiter_long_1997}, are more suitable when one considers cumulated numerical errors due to the deep-structured nature of RNNs\cite{bengio_learning_1994},
\begin{equation}
    \mathbf{H}^{(i)} = \overrightarrow{\mathbf{H}}^{(i)} + \overleftarrow{\mathbf{H}}^{(i)}.
    \label{eqn:group:concat}
\end{equation}
Hidden layers from the forward and backward RNNs are then merged into a single sequence, as described in Eq. \ref{eqn:group:concat}.
Finally, we obtain the sequence of chemical feature vectors of the $\alpha$-th atom in the given solvent with weighted summation of stacked rnn layers,
\begin{equation}
    \mathbf{P} = \sum_{i} c_{i} \mathbf{H}^{(i)}.
    \label{eqn:group:weighted_sum}
\end{equation}
The encoder function for solutes has an identical neural network architecture, which converts the pre-trained solute sequence $\mathbf{Y}$ into the feature sequence $\mathbf{Q}$.

We also consider the graph convolutional neural network (GCN), which is one of the most well-known algorithms in chemical applications of neural networks\cite{kipf_semi-supervised_2017,kearnes_molecular_2016}.
The GCN model represents the input molecule as a mathematical graph, instead of a simple sequence:
each node corresponds to the atom, and each edge in the adjacency matrix $\mathbf{A}$ involves connectivity (or existence of bonding) between atoms:
\begin{equation}
    \mathbf{H}^{(i + 1)} = \mathrm{GCN} ( \mathbf{H}^{(i)}, \mathbf{A} ).
\end{equation}
The role of adjacency matrix in the GCN constrains convolution filters to the node itself and its nearest neighbors. Eqn. \ref{eqn:group:skip_gcn} describes a more detailed mathematical expression of the skip-connected GCN\cite{kipf_semi-supervised_2017}
\begin{equation}
    \mathrm{GCN} (\mathbf{H}, \mathbf{A}) =
    \sigma (\mathbf{D}^{-1/2} \mathbf{A} \mathbf{D}^{-1/2} \mathbf{H} \mathbf{W}_{1} + \mathbf{H} \mathbf{W}_{2} + \mathbf{b}),
    \label{eqn:group:skip_gcn}
\end{equation}
where $\mathbf{D}$ is the degree matrix, $\mathbf{W}_{1}$ and $\mathbf{W}_{2}$ are convolution filters, $\mathbf{b}$ is the bias vector, and $\sigma$ denotes the activation function - we choose the hyperbolic tangent in the proposed model.
The GCN encoder also invloves stacked structure, and we can obtain the feature sequence for each molecule with the same manner as described in Eqn. \ref{eqn:group:weighted_sum}.

\bibliographystyle{achemso}
\bibliography{references}